# Weighted Tanimoto Coefficient for 3D Molecule Structure Similarity Measurement

Siti Asmah Bero, Azah Kamilah Muda\*, Yun-Huoy Choo, Noor Azilah Muda, Satrya Fajri Pratama

Computational Intelligence and Technologies (CIT) Research Group, Center of Advanced Computing and Technologies, Faculty of Information and Communication Technology, Universiti Teknikal Malaysia Melaka Hang Tuah Jaya, 76100 Durian Tunggal, Melaka, Malaysia

sitiasmah.bero@gmail.com, {azah, huoy, azilah}@utem.edu.my, satrya@student.utem.edu.my

**Abstract.** Similarity searching of molecular structure has been an important application in the Chemoinformatics, especially in drug discovery. Similarity searching is a common method used for identification of molecular structure. It involve three main principal component of similarity searching: structure representation; weighting scheme; and similarity coefficient. İn this paper, we introduces Weighted Tanimoto Coefficient based on weighted Euclidean distance in order to investigate the effect of weight function on the result for similarity searching. The Tanimoto coefficient is one of the popular similarity coefficients used to measure the similarity between pairs of the molecule. The most of research area found that the similarity searching is based on binary or fingerprint data. Meanwhile, we used non-binary data and was set amphetamine structure as a reference or targeted structure and the rest of the dataset becomes a database structure. Throughout this study, it showed that there is definitely gives a different result between a similarity searching with and without weight.

**Keywords:** Weighted Tanimoto Coefficient, 3D, molecule structure, similarity searching

## 1    Introduction

According to the Collins English Dictionary, words "similar" is defined as "showing resemblance in qualities, characteristics or appearance; alike but not identical". Similarity plays an important role in many aspects of life whereby, the definition of similarity itself is subjective to be interpreted and the result of similarity itself can be evaluated in many ways.

In chemistry, the similarity concept was applied through a long period of time and had become an important role in this domain [1]. Over the past few years, rapid technological advancements have triggered the development of structural similarity using computer tools [3-6]. Therefore, the Cheminformatics was established. The cheminformatics is defined as a domain that studies between chemistry and information technology. This domain is responsible to process, store, manipulation, and analysis of the chemical information [8].

Drug abuse among people worldwide is getting serious day by day. According to that, the knowledge discovery techniques on drug discovery have become an important domain for cheminformatics, especially on the identification of drugs. Amphetamine-type stimulants (ATS) is an example of drugs that was used widely around the world. ATS is a synthetic drug that comprised of amphetamine-group (primarily amphetamine, methamphetamine and methcathinone) and ecstasy-group substances (3,4-methylenedioxymethamphetamine (MDMA) and its analogs) [10]. Due to the existence of ATS drug, the abuse of drugs has become a global, harrowing social problem. It becomes a challenge due to the limitations of the current test kit to detect unfamiliar substances, besides sometimes it's also prone to false positive detection.

Meanwhile, forensic drug analysis deals with identification and quantification of illegal drugs that pass through a time consuming laboratory tests. Therefore, lots of identification of drugs was carried out using virtual screening (computational method) [11-12], for which the similarity searching approach is used [13-14]. Note that, the process of calculating the similarity searching of molecules involves both target structure and each of the structures in a database [9]. The target structure is referred as reference structure as it will be used as benchmark structure for the structures in the database.

İn this paper, we focus on similarity searching between Amphetamine (reference structure) with Amphetamine-type stimulants (ATS) drugs and Non-ATS drugs (database structure) by proposed a Weighted Tanimoto coefficient as the similarity measure. In this paper we used the non-binary data that is deduced from the three-dimensional (3D) structures of ATS drugs. A weighted Tanimoto Coefficient was used in this project for calculating the resemblance of similarity between the reference structure and database structure. The paper will be organized as follow: The introduction is at section 1, material and method will be at section 2, followed by result and discussion at section 3 and finally the conclusion will be at section 4.

## 2 Material and Method

### 2.1 Material

The dataset in this study was obtained from the *3D Exact Legendre* [17]. The dataset contains of feature extraction from 7212 3D molecular structure of ATS and non-ATS drugs; 3610 of ATS drug's structure and 3602 of non-ATS drug's struc-ture. İn this study, we used ATS drug as the traning set, whereby the non-ATS drug as a testing set.

**2.2 Method**

This study has computed the similarity searching between an amphetamine structure as a reference structure and the other molecule structures of dataset being the database structure. İn this study, we proposed a weighted Tanimoto Coefficient by adoption of Euclidean weight function to Tanimoto coefficient as the measure of similarity between those molecule structures. The Tanimoto coefficient is known as one of by far the most popular similarity measure for 2D fingerprint (binary data) molecule structure [19] as reported by many studies [2] [9] [13] [15], and this coefficient is found rarely used to non-binary data. Since the dataset of this study is in non-binary form, we had used the continuous form Tanimoto coefficient, together with the weight for each of molecules attribute in order to express the relative importance of each attribute.

**2.2.1 Tanimoto Coefficient**

Since the dataset for this study is in non-binary form, the continuous form of Tanimoto coefficient was chosen, which is suited for our dataset. $T_{R,D}$ is a similarity between the reference structure, R and database structure, D are represented by vector, $x$, of length $n$ with the $i$th property having the value $x_i$ [16]. The formula of continuos form of Tanimoto coefficient is given by the Eq. 1:

$$T_{R,D} = \frac{\sum_{i=1}^{n}(x_{iR}\ x_{iD})}{\sum_{i=1}^{n}(x_{iR})^2 + \sum_{i=1}^{n}(D_{iD})^2 + \sum_{i=1}^{n}(x_{iR}\ x_{iD})} \quad (1)$$

where,

$$\sum_{i=1}^{n}(x_{iR})^2 \quad (2)$$

And

$$\sum_{i=1}^{n}(x_{iD})^2 \quad (3)$$

And

$$\sum_{i=1}^{n}(x_{iR}\ x_{iD}) \quad (4)$$

The Eq. (2) and Eq.(3) indicates the sum of the squares of all elements in the reference data, R and database structure, D. Meanwhile, the Eq. (4) indicates the inner product (dot product) or the sum of the *products* for each element in R and D. The origin and the detail explanation can be found in [20].

There is lots of similarity coefficient been reported in the literature and can be categorized into three: distance coefficient, association coefficients and correlation coefficients [9]. The Tanimoto coefficient is lies under the association coefficient. The association coefficient is commonly used with binary data, that assigned a value that range from 1 (indicating the complete similarity) and 0 (indicating no similarity) [20][9].

According to [9], the association coefficients can be used for non-binary data, whereby the range of similarity might be different. The previous study done by a research group at the University of Sheffield, they had used the Tanimoto coefficient as the similarity measure in lots of their experiment on similarity searching of the molecular structure as presented in [2,4,7,14, 21-22]. The ease of implementation and the calculation that does not involve any square root, making it always had been a preferred similarity measure of molecular structure. A detail review on Tanimoto coefficient being an appropriate similarity measure for fingerprint can be found in [23].

**2.2.2 Euclidean Distance**

The Euclidean distance is known as one of an established distance metric in the field of Mathematics [24-25]. İn mathematics, this distance metric is known as the "ordinary" straight-line distance calculation between two points. By means, the Euclidean distance is used to measure distance (d) between point *x* and *y*, or from *y* to *x* is given by Eq. (6) [26]. To donate the distance between point x and y we can use the notation $d_{(x,y)}$ and written as [27]:

$$d^2_{(x,y)} = (x_1 - y_1)^2 + (x_2 - y_2)^2 \tag{5}$$

Then, the distance itself is the square root

$$d(x,y) = d(y,x) = \sqrt{(x_1 - y_1)^2 + (x_2 - y_2)^2 + \cdots + (x_n - y_n)^2} \tag{6}$$

This well-known distance coefficient can be used for *J* dimension, such as three-dimension, fourth-dimension and so on. The Eq. (7) below indicating the distance between two *J*-dimensional vectors *x* and *y* [27]:

$$d(x,y) = \sqrt{\sum_{i=1}^{J}(x_i - y_i)^2} \tag{7}$$

**2.2.3 Weighted Tanimoto Coefficient**

There are many way to compute the weight function of attributes. The main reason why the weight is assigned to each attributes, because some attributes may be more important than others and in some case, weights can be defined to express the relative importance of the attributes [27]. One of the novel method for multiple attribute decision making (MADM) using weighted Euclidean distance, known as

Weighted Euclidean Distance Based Approach (WEBDA) has been reported by [28].

In this paper, the calculation of the attributes weight function is inspired by weighted euclidean distance [27] that defined the weight as the inverse of the *j*-th variance. The procedure of the Weighted Tanimoto coefficient is given as follow:

*Step 1: Standardization:*
The standardization of attribute data is used to balance out the contributions. The important process of standardization is used to transform the variables so they all have the same variance of 1 and mean of zero [27-28]. The standardized attribute data is sometimes called as standard score or z-score. The standardized attribute data is given as follows:

$$Z_{ij} = \frac{x_{ij} - \mu_j}{\sigma_j} \quad (8)$$

$$\mu_j = \frac{1}{m}\sum_{i=1}^{m} x_{ij} \quad (9)$$

$$\sigma_j = \sqrt{\frac{\sum_{i=1}^{m}(x_{ij}-\mu_j)^2}{m}} \quad (10)$$

The $Z_{ij}$ is the z-score. Where, $x_{ij}$ is the value of each element, $\mu_j$ is the expected value or population mean of $j^{th}$ attribute and $\sigma_j$ is the standard deviation of $j^{th}$ attribute.

*Step 2: Weighted Tanimoto Coefficient:*
Based on weighted Euclidean distance [27], the weight of attributes can be calculated as follow:

$$W_j = \frac{1}{\sigma^2_j} \quad (11)$$

Where the weight = $W_j$ is the inverse of the *j*-th variance. The $W_j$ is considered as the weight that attached to the *j*-th variable.

Then, the Weighted Tanimoto coefficient is written as follow:

$$WT_{R,D} = \frac{\sum_{i=1}^{n}(W_j x_{iR}\ W_j x_{iD})}{\sum_{i=1}^{n}(W_j x_{iR})^2 + \sum_{i=1}^{n}(W_j x_{iD})^2 + \sum_{i=1}^{n}(W_j x_{iR}\ W_j x_{iD})} \quad (12)$$

Where, at first, each attribute is multiplied by their corresponding weights and then the similarity score for reference structure and database structure is computed. Next process, the similarity searching was solved by using Weighted Tanimoto coefficient.

## 3 Simulation Result and Discussion

### 3.1 Simulation Result

İn order to calculate the similarity score of the dataset, the dataset need to follow these step:

The step of calculating similarity score with weight:
1. Firstly, the dataset needs to be standardized by using the Eq. (8) to Eq. (10).
2. After that, the weight for each attribute of standardized data was calculated by using the Eq. (11) and only then,
3. The similarity searching score was computed by using the Eq. (12).

Meanwhile, to calculate the similarity score without weight function, we just undergo two main steps:
1. Firstly, the dataset needs to be standardized by using the Eq. (8) to Eq. (10) and,
2. Secondly, the similarity searching score was computed by using the Eq. (1).

The result of similarity searching with weight was ranked in decreasing order as shown in Table 2. Meanwhile, the Table 3 shows comparison between the Amphetamines (reference structure) with five different drug's molecule structure with their corresponding similarity class. Each similarity score obtained was classified into six different class of similarity: 1- Very high, 2- High, 3- Medium, 4- Low, 5- Very low and 6- None (indicating no similarity) [29]. The interpretation of class similarity is given in Table 1.

Table 1. The interpretation of the correlation class similarity for similarity score

| Correlation of class similarity | Dependance between variables |
|---|---|
| 1 | Absolute |
| 0.9-1 | Very High |
| 0.7-0.9 | High |
| 0.4-0.7 | Medium |
| 0.2-0.4 | Low |
| 0-0.2 | Very Low |
| 0 | None |

Table 2: Summarization of Similarity score of datasets with their corresponding class of similarity

| Class of Drugs | Molecule ID | Similarity Score | Class of Similarity |
|---|---|---|---|
| ATS | pk2006 | 1 | ABSOLUTE |
| ATS | pk457 | 0.9576 | VERY HIGH |
| ATS | pk10495 | 0.8954 | HIGH |
| : | : | : | : |
| NATS | NATS2345089 | 0.7099 | MEDIUM |
| ATS | pk6828 | 0.7098 | MEDIUM |
| ATS | pk10779 | 0.7097 | MEDIUM |
| : | : | : | : |
| NATS | NATS295209 | 0.4048 | LOW |
| ATS | pk6534 | 0.4047 | LOW |
| NATS | NATS4028903 | 0.4045 | LOW |
| : | : | : | : |
| ATS | pk5394 | 0.2029 | VERY LOW |
| NATS | NATS6317334 | 0.2026 | VERY LOW |
| ATS | pk1256 | 0.2024 | VERY LOW |
| : | : | : | : |
| NATS | NATS2289131 | 0.0112 | NONE |
| NATS | NATS347078 | 0.0013 | NONE |

Table 3: Comparison between the Amphetamine (reference structure) and other drug's molecule structure with their corresponding similarity class

| Reference Structure | Database structure | Similarity score (without weight) | Similarity score (with weight) | Class of Similarity |
|---|---|---|---|---|
| 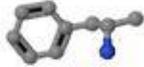 Figure 1. Amphetamine | 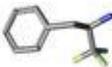 Figure 3. pk457 | -0.1399 | 0.9576 | Very high |
| | 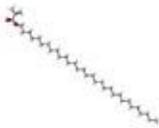 Figure 8. NATS2289131 | 0.0772 | 0.0112 | None |

## 3.2 Discussion

The amphetamines have been chosen as reference structure as it is basic structure of the ATS drugs. The experiment was conducted in order to study the effect of weight function on the result of similarity searching for ATS drugs. The Weighted Tanimoto Coefficient proposed in this study is used to calculate the similarity searching of the molecule structure. The intergration of weight function into Tanimoto Coefficient clearly shows improvement in similarity score of molecular structure as shown in Table 3.

For example, the result of similarity score for the molecule pk457, represent of ATS drug and NATS2289131, represent of non-ATS drug without weight is -0.1399 and 0.0772 respectively. After implementing the weight, the similarity score shows major differences for both molecules. This situation is related to the relationship between the properties of clusters; inter-class and the intra-class concept. The *inter-* is referred as between or among group, while, the *intra-* is referred as on the inside, within group [30]. A good clustering is said to produce a high quality clusters with a high intra-class similarity and low inter-class similarity.

ATS drugs have three main groups: Amphetamine, methamphetamine, and ecstasy. We can say that the structure that belongs to the same group of amphetamine, methamphetamine or ecstasy as intra-class. Thus, the inter-class is referred to similarity searching among the group of ATS drugs. The Figure 3 and 4 have slightly highest similarity score among the others due to their structure is almost the same with the amphetamine as they had less substructure molecule attached to them. In contrast with the Figure 7 and 8, the weight function had affected their similarity score and if we look with the bare eyes, both structure of Figure 7 and 8 is totally had a different structure with amphetamine, this is show by it similarity score that has a very low value of similarity score.

## 4    Conclusion

In this study we have proposed a non-binary similarity measure for the molecular structure by integrating weighted Euclidean distance to Tanimoto coefficient. The purpose is to enhance the effectiveness of similarity searching. In this study, the weight function is used to express the relative importance of attributes that lead to the differentiation value between the inter-class and intra-class. From the simulation result, its shows that the weight will affects the similarity score within the same class (intra-class) and it make a large differentiation (decrease the similarity score) between the molecules of other class (inter-class). The degree of similarity within a molecule also related to their substructure molecule that attached to them; the more substructure of molecule attached, the less similar it to reference structures. In the other hand this proved that, if the molecules share more similar structure, it tends to get a higher Tanimoto coefficient score and tends to agree with the *similar property principle*: the structurally similar molecules tend to have similar properties [15][31]. For the future work, the study is intended to do a clustering accuracy of the result obtained.

**References**

1. Rouvray, D.H. (1990) The evolution of the concept of molecular similarity, in Concepts and Applications of Molecular Similarity (Johnson, M. A., and Maggiora, G. M., Eds.), pp 15–42, John Wiley, Chichester.
2. Willett, P.: Similarity Searching Using 2D Structural Fingerprints. Methods Mol. Biol. 672, 133–158 (2011).
3. Eckert H, Bajorath J. Molecular similarity analysis in virtual screening: foundations, limitations and novel approaches. Drug discovery today Mar 31;12(5):225-33 (2007).
4. Willett, P.: Similarity methods in chemoinformatics. Annu. Rev. Inf. Sci. Technol. 43, 1–117 (2009).
5. Maggiora, G.M., Shanmugasundaram, V.: Molecular Similarity Measures. In: Bajorath, J. (ed.) Chemoinformatics and Computational Chemical Biology. pp. 39–100. Humana Press, Totowa, NJ (2011).
6. D. Stumpfe, J. Bajorath, WIRES Comp. Mol. Sci. 2011, 1, 260.
7. Willett, P., Barnard, J.M., Downs, G.M.: Chemical Similarity Searching. J. Chem. Inf. Model. 38, 983–996 (1998).
8. Maccuish, J.D., Maccuish, N.E.: Chemoinformatics applications of cluster analysis. Wiley Interdiscip. Rev. Comput. Mol. Sci. 4, 34–48 (2014).
9. Holliday: Grouping of Coefficients for the Calculation of Inter-Molecular Similarity and Dissimilarity using 2D Fragment Bit-Strings. Comb. Chem. High Throughput Screen. 5, 155−166 (2002).
10. United Nations Office on Drugs and Crime: World Drug Report 2010. (2010).
11. Koeppen, H. Virtual Screening?What Does It give Us? Curr. Opin. Drug Discovery Dev. 2009, 12, 397−407.
12. Willett, P. Similarity-Based Virtual Screening Using 2D Fingerprints. Drug Discovery Today 2006, 11, 1046−1053.
13. Stumpfe, D.; Bajorath, J. Similarity Searching. Wiley Interdiscip. Rev.: Comput. Mol. Sci. 2011, 1, 260−282.
14. Willett, P.: The calculation of molecular structural similarity: Principles and practice. Mol. Inform. 33, 403–413 (2014).
15. Todeschini, R., Consonni, V., Xiang, H., Holliday, J., Buscema, M., Willett, P.: Similarity coefficients for binary chemoinformatics data: Overview and extended comparison using simulated and real data sets. J. Chem. Inf. Model. 52, 2884–2901 (2012).
16. Sheridan, R. P. Chemical similarity searches: when is complexity justified? Expert Opin. Drug Discov. 2007, 2, 423−430.
17. Pratama. S.F, Muda. A.K, Choo. Y.H, Abraham. A, "Exact Computation of 3D Geometric Moment Invariants for ATS Drugs Identification.," *Proceedings of the 6th International Conference on Innovations in Bio-Inspired Computing and Applications.*, vol.424, pp. 347-358, 2016.

18. Maggiora, G.M., Vogt, M., Stumpfe, D., Bajorath, J.J.: Molecular similarity in medicinal chemistry. J. Med. Chem. 57, 3186–3204 (2013).
19. Ellis, D., Furner-Hines, J., Willett, P.: Measuring the degree of similarity between objects in text retrieval systems. Inf. Manag. 3.2, 128–149 (1993).
20. Bajusz, D., Rácz, A., Héberger, K.: Why is Tanimoto index an appropriate choice for fingerprint-based similarity calculations？ J. Cheminform. 1–13 (2015).
21. P. Willett, V. Winterman, and D. Bawden.: Implementation of nearest-neighbor searching in an online chemical structure search system,. *J. Chem. Inf. Model.*, vol. 26, pp. 36–41(1986).
22. P. Willett, V. Wintermann.: A Comparison of Some Measures for the Determination of Inter-Molecular Structural Similarity Measures of Inter-Molecular Structural Similarity. *Quant. Struct. Relationships*, vol. 25, pp. 18–25 (1986).
23. Dattorro, J. Convex optimization & Euclidean distance geometry, Meboo Publishing, California(2008)
24. Gower, J.C.: Euclidean distance geometry, http://convexoptimization.com/TOOLS/Gower2.pdf, (1982).
25. Euclidean distance , https://en.wikipedia.org/wiki/Euclidean_distance
26. Marques Pereira, R.A., Ribeiro, R.A.: Aggregation with generalized mixture operators using weighting functions. Fuzzy Sets Syst. 137, 43–58 (2003)
27. Greenacre, M., Primicerio, R.: Measures of distance between samples: Euclidean. Multivar. Anal. Ecol. Data. 47–59 (2013).
28. Rao, R.V., Singh, D.: Weighted Euclidean distance based approach as a multiple attribute decision making method for plant or facility layout design selection. Int. J. Ind. Eng. Comput. 3, 365–382 (2012).
29. Kl, M.: Evaluation of ranking similarity in ordinal ranking problems. 119–128.
30. Writing Explained, http://writingexplained.org/inter-vs-intra-difference
31. Shin, W.-H., Zhu, X., Bures, M., Kihara, D.: Three-Dimensional Compound Comparison Methods and Their Application in Drug Discovery. Molecules. 20, 12841–12862 (2015).